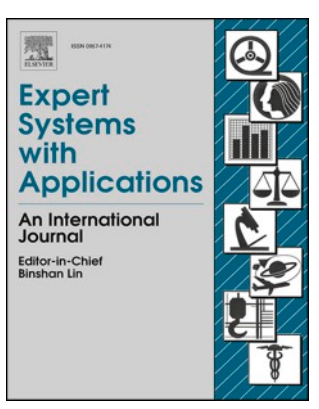

# Integrated multivariate segmentation tree for heterogeneous credit data analysis in small- and medium-sized enterprises

Lu Han [a], Xiuying Wang [b,*]

[a] School of Management Science and Engineering, Central University of Finance and Economics, Beijing 100081, China
[b] School of Computer Science, University of Sydney, NSW 2006, Australia

ARTICLE INFO

ABSTRACT

Traditional decision tree models, which rely exclusively on numerical variables, often face challenges in handling high-dimensional data and are limited in their ability to incorporate textual information effectively. To address these limitations, we propose the integrated multivariate segmentation tree (IMST), a comprehensive framework designed to improve credit evaluation for small- and medium-sized enterprises (SMEs) by integrating financial data with textual sources. This method comprises three core stages: (1) transforming textual data into numerical matrices through matrix factorization, (2) selecting salient financial features using Lasso regression, and (3) constructing a multivariate segmentation tree based on either the Gini index or entropy, with weakest-link pruning applied to control model complexity. Experimental results based on a dataset of 1,428 Chinese SMEs demonstrated that IMST achieved an accuracy rate of 88.9%, surpassing both baseline decision trees (87.4%) and conventional models such as support vector machines and neural networks. Furthermore, the proposed model demonstrated superior interpretability and computational efficiency, featuring a more streamlined architecture and improved risk detection capabilities.

## 1. Introduction

In the rapidly evolving landscape of financial services, credit evaluation of small- and medium-sized enterprises (SMEs) has become critical for banks. Traditional models that rely solely on numerical data—such as profit, liabilities, and assets—are increasingly supplemented by unconventional text-based data sources. This integration of structured numerical data and unstructured textual information marks a paradigm shift in credit evaluation, offering considerable opportunities to improve predictive accuracy, reduce bias, and promote financial inclusion (Jiang, Yin, Tang, & Wang, 2023; Lee, Yang, & Anderson, 2024). Strategic combination of these data types addresses the limitations of conventional approaches and provides valuable insights into borrower behavior that numerical indicators alone cannot capture (Gutierrez-Gomez, Petry, & Khadraoui, 2020). However, effective models capable of handling heterogeneous data in credit evaluation tasks are lacking.

Decision tree models are widely used in credit assessment owing to their distinct characteristics; however, they have certain limitations (Martin, 2013; Zhang & Yu, 2024). These models generate clear and interpretable decision rules, facilitating stakeholder understanding and credit decision validation (Lee, Yen, Jiang, Chen, & Chang, 2025). Moreover, decision trees accommodate both categorical and numerical data, thereby reducing the need for extensive preprocessing. They also typically exhibit fast training speeds, enabling rapid model updates—an essential capability for real-time credit scoring in dynamic environments (Yang, Yan, Qiao, Wang, & Qian, 2025). However, decision trees encounter challenges when handling high-dimensional continuous variables. Discretization of continuous features may lead to loss of fine-grained information and increased model complexity (Tao et al., 2021). In summary, the development of decision trees with multivariate splitting mechanisms will not only improve model efficiency but also enhance their applicability in credit evaluation.

To address the challenge of constructing interpretable models using heterogeneous data, we developed a novel approach: integrated multivariate segmentation tree (IMST). This model effectively integrated financial statements and short textual descriptions from loan audits for credit evaluation. The main contributions of this study could be summarized as follows:

(1) This study introduced an innovative text compression method that converts textual records into numerical matrices, providing a novel data representation and processing approach. The main idea here was to leverage mathematical matrices to encapsulate essential information from text data while reducing the data size to enhance storage and transmission efficiency. This technique ensured compatibility with diverse computational algorithms and machine learning models operating on numerical inputs by transforming textual information into numerical form.

* Corresponding author.
*E-mail addresses:* hanluivy@126.com (L. Han), xiu.wang@sydney.edu (X. Wang).

(2) An IMST model that integrates multiple variables into a unified framework using Lasso regression was developed. This model leveraged the interdependencies among various features to generate more informed splitting decisions at each node. In contrast to conventional decision trees, which evaluate one variable at a time, IMST considers multiple variables simultaneously, resulting in more accurate and nuanced predictions.

(3) To validate the effectiveness of the proposed model, extensive experiments were conducted using several benchmark models. The results consistently revealed that IMST outperformed competing models in terms of both computational speed and predictive accuracy.

The remainder of this paper is organized as follows: Section 2 reviews the related work, focusing on two key areas—classification models for heterogeneous data and decision trees; Section 3 describes the IMST model in detail; Section 4 presents and analyzes the experimental results; Section 5 discusses the applicability of the proposed method and its limitations; and Section 6 outlines the conclusions and directions for future research.

## 2. Related work

Heterogeneous data refer to datasets containing features of various formats, such as numerical, categorical, and textual data. Compared with traditional unimodal or homogeneous datasets, heterogeneous data are characterized by their inherent diversity (Browne & McNicholas, 2012). This diversity manifests not only in the data values but also in the heterogeneous nature of the data, introducing unique challenges for classification tasks (Lu, Chen, Wang, & Lu, 2016; Stahlschmidt, Ulfenborg, & Synnergren, 2022). Conventional classification algorithms often struggle to effectively integrate heterogeneous information from different modalities (Xu & Tian, 2025). They encounter difficulties in modeling interactions among heterogeneous features, capturing considerable variations caused by anomalies and handling complex, analytically intractable posterior inference tasks (Li et al., 2022; Li & Tang, 2024). When the heterogeneity within the data is treated as noise or contamination, constructing reliable classification models becomes even more challenging (Gao, Li, Chen, & Zhang, 2020). Therefore, an effective integration of these heterogeneous information sources to achieve accurate classification is currently a key research focus (Zhao, Zhang, & Geng, 2024).

As a classic supervised learning algorithm, decision trees have been extensively applied to classification and regression tasks because of their interpretability and ease of understanding (Han, Li, & Su, 2019; Mienye & Jere, 2024). However, traditional decision tree algorithms encounter considerable challenges when handling heterogeneous data (Blockeel, Devos, Frénay, Nanfack, & Nijssen, 2023).

The core of decision trees lies in selecting the optimal feature for node splitting. Traditional decision tree algorithms, such as ID3, C4.5, and CART, use different splitting criteria—such as information gain, gain ratio, and the Gini index—for categorical and numerical data (Abolhosseini, Khorashadizadeh, Chahkandi, & Golalizadeh, 2024; Charbuty & Abdulazeez, 2021; Han, Li, & Su, 2019). However, when datasets contain mixed variable types, establishing a unified measure of feature importance is a key challenge. Recently, new splitting criteria have been proposed to better accommodate heterogeneous data. One approach is based on similarity measures. Zhang and Jiang (2012) introduced a similarity-based splitting criterion that leverages the similarity among class labels to guide the splitting process to generate more homogeneous subsets. This method can handle both discrete and continuous attributes. For ordinal categorical data, some studies have focused on developing splitting criteria that capture the inherent order among categories. Researchers have reviewed and compared several criteria, including ordinal Gini, weighted information gain, and ranking impurity, which enable the splitting process to better utilize ordinal information and thereby improve classification performance (Khalaf, Garcia, & Ben Ishak, 2025). Moreover, with advances in natural language processing, recent studies have explored the use of large language models (LLMs) to assist decision tree splitting. Carrasco, Urrutia, and Abeliuk (2025) proposed zero-shot decision tree construction that leverages the pretrained knowledge of LLMs to suggest potential split points based on the name, type, and description of each feature. This method estimates the label distribution of the resulting subsets and selects the optimal split by minimizing Gini impurity. Furthermore, it enables the construction of decision trees for heterogeneous data without requiring training data, offering a novel solution for scenarios with limited data availability.

To further improve the performance of decision trees on heterogeneous data, researchers have investigated hybrid strategies that integrate decision trees with other machine learning models (Gupta & Kishan, 2025; Thirunavukkarasu & Jamal, 2025) as well as ensemble learning approaches (Khan, Chaudhari, & Chandra, 2024; Hsu, Tsao, Chang, & Chang, 2021).

One approach involved combining decision trees with other models in either a sequential or parallel manner to leverage their respective strengths in handling different aspects of heterogeneous data (Li, Herrera-Viedma, Kou, & Morente-Molinera, 2023). Scholars have proposed a hybrid model that integrates support vector machines (SVMs) with decision trees for kidney disease detection (Thirunavukkarasu & Jamal, 2025). In this model, SVM is first used for processing numerical features, and its prediction results are then used as input features for a decision tree to handle image data, thereby enabling effective classification of heterogeneous data. Arifuzzaman, Hasan, Toma, Hassan, and Paul (2023) introduced a decision tree-based deep neural network, which uses a deep neural network to learn complex feature representations, followed by a decision tree structure for classification. This model is particularly suitable for nonlinearly separable data. A few researchers have proposed hybrid decision tree algorithms for regression analysis involving both numerical and categorical data (Kim & Hong, 2017; Tao et al., 2021; Zhou, Zhang, Zhou, Guo, & Ma, 2021). Their approach first uses a decision tree to estimate the effect of categorical variables on the target variable and then applies other regression algorithms to predict the residuals based on numerical features.

Ensemble learning is another approach to handling heterogeneous data. The gradient-boosted decision tree (GBDT) is the most well-known method in this category; it iteratively builds weak learners, typically decision trees, by focusing on the errors made by previous learners at each step, thereby gradually improving model performance (Emami & Martínez-Muñoz, 2025). GBDT and its variants, such as XGBoost and LightGBM, have demonstrated strong capabilities in handling heterogeneous data (Airlangga & Liu, 2025). Furthermore, random forests (Paim & Enembreck, 2025) and other ensemble learning methods (Xia, Wang, Lan, Liu, & Wu, 2025), including bagging, boosting, and stacking, have been applied to improve the performance of decision trees on heterogeneous data.

While ensemble methods are widely acknowledged for their superior predictive performance, advancing single decision tree models remains critically important in domains requiring high interpretability, decision transparency, and computational efficiency—such as credit scoring. In such contexts, a more powerful and robust single decision tree can offer greater practical utility compared to ensemble-based approaches. Significant progress has been made in developing decision tree methodologies for heterogeneous data. Enhanced splitting criteria, along with the emergence of hybrid and ensemble techniques, have substantially improved the performance and applicability of decision trees in handling complex and diverse datasets. A summary of selected studies on single decision tree models for heterogeneous data is presented in Table 1. However, persistent challenges include the effective integration of heterogeneous feature types and the design of efficient, scalable hybrid and ensemble architectures. As data complexity continues to grow, research focused on enhancing decision trees for heterogeneous data will remain a vital direction and is expected to play an increasingly prominent role across a wide range of real-world applications.

**Table 1**
Summary of selected studies on single decision tree models.

| Works | Method | Data, features, and performance |
|---|---|---|
| Tao et al., 2021 | Introduces generalized piecewise linear decision tree (GPWL-DT), a novel decision tree model integrating continuity and piecewise linearity into the decision-making process, effectively overcoming the inherent discontinuity of conventional piecewise constant decision trees such as CART and C4.5. | 16 datasets from UCI Repository, such as Bodyfat, Spacega, Abalone, and Cpusmall |
| Zhou, Zhang, Zhou, Guo, & Ma, 2021 | Proposes FWDT, a hybrid feature selection method that integrates an enhanced relief-based filtering stage with an embedded decision tree framework. By utilizing the median of feature weights as an adaptive threshold, the method effectively performs preliminary filtering of irrelevant features, thereby improving the relevance and computational efficiency of the subsequent feature selection. | 12 datasets from UCI Machine Learning Repository, such as Spect, Ecoli, SPECTF, and Horse-Colic |
| Li, Herrera-Viedma, Kou, & Morente-Molinera, 2023 | Introduces a Z-number-valued rule-based decision tree, which extends the conventional fuzzy rule-based decision tree by integrating Z-numbers to manage uncertain and partially reliable information more effectively. | 13 datasets from UCI Machine Learning Repository, such as Authorship, Credit, Trial, and Iris |
| Abolhosseini, Khorashadizadeh, Chahkandi, & Golalizadeh, 2024 | Proposes modified ID3 decision tree algorithm, referred to as cumulative residual entropy-based decision tree, which uses cumulative residual entropy as the splitting criterion in place of traditional Shannon entropy. | 10 real-world datasets: student performance, California housing, red wine quality, concrete compressive strength, abalone age prediction, fish market, energy efficiency, forest fires, births and deaths in New Zealand, and insurance agency performance |
| Khalaf, Garcia, & Ben Ishak, 2025 | Introduces local adaptive streaming tree (LAST), a decision tree model for high-speed data streams. It features an adaptive splitting mechanism that creates new branches when performance degrades, allowing dynamic adaptation to changes in the data stream and the tree. | Datasets from UCI Machine Learning Repository |

## 3. Integrated multivariate segmentation tree

This section presents IMST, a method designed to handle mixed-type data. The proposed approach consisted of three main stages: first, short text records were transformed into a latent matrix; second, feature selection was conducted for multidimensional numerical variables; and third, the multivariate segmentation tree was constructed. Notations for the variables used in our model are listed in Table 2.

### 3.1. Construction of latent matrix for textual data

Herein, we propose a matrix factorization method for constructing a latent matrix for textual data. D was assumed to be the $n \times d$ document-term matrix, where n documents were defined with a lexicon of size d. Then, the non-negative matrix was decomposed into two matrices U and V, which minimized the following objective function:

$$J = \frac{1}{2}\left\| D - UV^T \right\|_F^2. \tag{1}$$

Here, $\|\cdot\|_F^2$ represents the squared Frobenius norm, $\|\cdot\|_1$ denotes the $L_1-$norm, U is the $n \times k$ non-negative matrix, and V denotes the $d \times k$ non-negative matrix. The value of k is the dimensionality of the embedding. Matrix U provides the new k-dimensional coordinates of the rows of D in the transformed basis system, and matrix V provides the basis vectors in terms of the original lexicon.

The objective function J was then expressed as follows:

$$\begin{aligned} J &= \frac{1}{2} tr\left[\left(D - UV^T\right)\left(D - UV^T\right)^T\right] \\ &= \frac{1}{2}\left[tr\left(DD^T\right) - tr\left(DVU^T\right) - tr\left(UV^T D^T\right)^T + tr\left(UV^T VU^T\right)\right]. \end{aligned} \tag{2}$$

With respect to matrices $U = [u_{ij}]$ and $V = [v_{ij}]$, and constraints $u_{ij} \geq 0$ and $v_{ij} \geq 0$, we set $P_\alpha = [\alpha_{ij}]_{n\times k}$ and $P_\beta = [\beta_{ij}]_{d\times k}$ to be matrices with the same dimensions as U and V. The elements of matrices $P_\alpha = [\alpha_{ij}]_{n\times k}$ and $P_\beta = [\beta_{ij}]_{d\times k}$ were the corresponding Lagrange multipliers for the non-negativity conditions on the different elements of U and V. Then, $tr(P_\alpha U^T)$ was equal to $\sum_{i,j}\alpha_{ij}u_{ij}$, and $tr(P_\beta V^T)$ was equal to $\sum_{i,j}\beta_{ij}v_{ij}$, respectively. Thus, the augmented objective function with constraint penalties can be expressed as follows:

$$L = J + tr\left(P_\alpha U^T\right) + tr\left(P_\beta V^T\right). \tag{3}$$

**Table 2**
Notations for variables used in the proposed model.

| Notation | Variable |
|---|---|
| $D$ | Document text matrix |
| $V$ | Document latent matrix after decomposed transformation with elements as $v_{ij}$ |
| $U$ | Document dimensional coordinates after decomposed transformation with elements as $u_{ij}$ |
| $P$ | Corresponding Lagrange multipliers of document transformation, with elements as $P_\alpha = [\alpha_{ij}]_{n\times k}$ and $P_\beta = [\beta_{ij}]_{d\times k}$. |
| $F$ | Numerical data matrix |
| $\beta$ | Coefficients of Lasso regression |
| $\hat{\beta}$ | Coefficients in Lagrangian form of Lasso regression |
| $\delta$ | Penalty parameter in Lasso regression |
| N | Categorical data matrix |
| S | Set on each class of categorical variables, with elements as $S_i$ |
| $p_j$ | Class distribution of each category |
| $\|T\|$ | Number of nodes in a tree |
| $N_m$ | Number of subtrees in a decision tree |
| $Q_m$ | Total error of a tree |
| $C_\alpha$ | Total complexity of a tree |

To optimize the problem in Eq. (3), the matrix calculus on the trace-based objective function yielded the following:

$$\begin{aligned}\frac{\partial L}{\partial U} &= -DV + UV^TV + P_\alpha = 0 \\ \frac{\partial L}{\partial V} &= -D^TU + VU^TU + P_\beta = 0.\end{aligned} \tag{4}$$

Using the Kuhn–Tucker conditions $\alpha_{ij}u_{ij} = 0$ and $\beta_{ij}v_{ij} = 0$, we rewrote the (i, j)-th pair of constraints as follows:

$$\begin{aligned}(DV)_{ij}u_{ij} - \left(UV^TV\right)_{ij}u_{ij} = 0 \\ \left(D^TU\right)_{ij}v_{ij} - \left(VU^TU\right)_{ij}v_{ij} = 0\end{aligned} \tag{5}$$

These conditions were independent of $P_\alpha$ and $P_\beta$, and the equations were solved using the following iterative methods for multiplicative updates for $u_{ij}$ and $v_{ij}$:

$$\begin{aligned}u_{ij} &= \frac{(DV)_{ij}u_{ij}}{\left(UV^TV\right)_{ij}}\forall i \in \{1, \cdots, n\}, \forall j \in \{1, \cdots, k\} \\ v_{ij} &= \frac{\left(D^TU\right)_{ij}v_{ij}}{\left(VU^TU\right)_{ij}}\forall i \in \{1, \cdots, d\}, \forall j \in \{1, \cdots, k\}.\end{aligned} \tag{6}$$

After the iterations were executed to convergence, the columns of V provided the basis for discovering document latent; and the columns of U, a basis corresponding to the latent.

### 3.2. Feature combination for numerical data

Lasso regression can shrink features by imposing a size penalty. Each coefficient is translates by a constant factor $\lambda$, which can be defined in Lagrangian form as follows:

$$\widehat{\beta}^{lasso} = \underset{\beta}{\operatorname{argmin}}\left\{\frac{1}{2}\sum_{i=1}^{N}\left(y_i - \beta_0 - \sum_{j=1}^{p}x_{ij}\beta_j\right)^2 + \delta\sum_{j=1}^{p}\left|\beta_j\right|\right\}. \tag{7}$$

The use of $L_1$ penalty $\delta\sum_{j=1}^{p}\left|\beta_j\right|$ resulted in a subset of the solution coefficients $\widehat{\beta}_j$ to be exactly zero for a sufficiently large value of the tuning parameter $\delta$. An efficient procedure for computing the Lasso solution for all $\delta$ is presented as Algorithm 1.

***Algorithm 1***. Feature selection using incremental forward stagewise regression

**Inputs:**
$X$: $n \times p$ matrix of predictors, where $n$ denotes the number of samples and $p$ indicates the number of features.
$y$: n-dimensional vector of labels.
**Outputs:**
$\beta$: p-dimensional vector of coefficients after feature selection.
**Step 1:** Initialize
- Set $\beta_1, \beta_2, \cdots\beta_p = 0$
- Set residual $r = y$ and $\varepsilon$=0.01

**Step 2:** While $\delta_j \leq threshold$
$c_j = x_j^Tr$;
$j^* = \operatorname{argmax}_j\left|c_j\right|$;
$\delta_j = \varepsilon\cdot sign\langle x_j, r\rangle$;
$\beta_{j^*} = \beta_{j^*} + \delta_j$;
$r = y - X\beta_{j^*}$;
j++;
End while
**Step 3:** Return $\{\beta_1^*, \beta_2^*, \cdots, \beta_k^*\}$

Using the predictors $\{\beta_1^*, \beta_2^*, \cdots, \beta_k^*\}$, we computed the new dataset F as follows:

$$f = \beta^*X. \tag{8}$$

Then, set F was used for building IMST. As Lasso regression shrank features, the number of branches of the decision tree decreased considerably; meanwhile, the errors of the method were reduced by processing the feature values.

### 3.3. Integrated multivariate segmentation tree

Subsequently, both the Gini index and entropy were used for constructing the multivariate segmentation tree (Mantas & Abellán, 2014; Mantas, Abellán, & Castellano, 2016; Moral-García, Mantas, Castellano, Benítez, & Abellán, 2020).

Gini index: $G(S) = 1 - \sum_{j=1}^{k}p_j^2$, where $G(S)$ is for a set of S on the class distribution $p_1, p_2, \cdots, p_k$. The overall Gini index for an r-way split of set S into sets $S_1, \cdots, S_r$ was quantified as the weighted average of the Gini index values G($S_i$) of each $S_i$, and the weight of $S_i$ was $|S_i|$. Therefore, the split with the lowest Gini index was calculated as follows:

$$Gini_Split(S \rightarrow S_1, S_2, \cdots S_r) = \sum_{i=1}^{r}\frac{|S_i|}{|S|}G(S_i). \tag{9}$$

Entropy measures for a set S were calculated according to $E(S) = -\sum_{j=1}^{k}p_j\log_2\left(p_j\right)$ on the class distribution $p_1, \cdots, p_k$ of the data points in each node. Furthermore, the overall entropy for an r-way split of set S into sets $S_1, \cdots, S_r$ was computed as the weighted average of $E(S_i)$ of each $S_i$, and the weight of $S_i$ was $|S_i|$. Therefore, the split with the lowest entropy was calculated as follows:

$$Entropy_Split(S \rightarrow S_1, S_2, \cdots S_r) = \sum_{i=1}^{r}\frac{|S_i|}{|S|}E(S_i). \tag{10}$$

The stopping criterion for the tree's growth was to explicitly penalize model complexity using weakest-link pruning. The cost of a tree was defined by a weighted sum of its error and complexity. We defined the terminal nodes as $m$, and $|T|$ represented the number of terminal nodes in the tree. Furthermore, $c_m$ indicated the predictors of the tree, and $Q_m$ denoted the total error of the tree, which was calculated as follows:

$$Q_m(T) = \frac{1}{N_m}\sum_{x_i \in T}\left(y_i - \widehat{c}_m\right)^2. \tag{11}$$

Next, the cost of model complexity is defined as follows:

$$C_\alpha(T) = \sum_{m=1}^{|T|}N_mQ_m(T) + \alpha|T|. \tag{12}$$

For each $\alpha$, the subtree $T_\alpha \subseteq T_0$, to minimize $C_\alpha(T)$, the tuning parameter $\alpha \geq 0$ governed the trade-off between tree size and its goodness of fit to the data. For each $\alpha$, we chose the smallest subtree that minimized $C_\alpha(T)$. We successively collapsed the internal node that produced the smallest per-node increase in $\sum_mN_mQ_m(T)$ and continued until the single-node tree was produced, yielding a finite sequence of subtrees that could be used to find the minimum. Next, $\alpha$ was estimated using cross validation; we chose the value $\widehat{\alpha}$ to minimize the cross-validated sum of squares. The IMST is summarized in Algorithm 2.

***Algorithm 2***. Integrated Multivariate Segmentation Tree

**Inputs:** $n \times m$ dataset S = {F, U}
**Outputs:** IMST.
**Step 1:** Create a root node containing S.
**Step 2:** While there exists an eligible node and $\alpha \geq 0$
$Gini_Split(S \rightarrow S_1, S_2, \cdots S_r) = \sum_{i=1}^{r}\frac{|S_i|}{|S|}G(S_i)$ $Entropy_Split(S \rightarrow S_1, S_2, \cdots S_r) = \sum_{i=1}^{r}\frac{|S_i|}{|S|}E(S_i)$ $Q_m(T) = \frac{1}{N_m}\sum_{x_i \in T}\left(y_i - \widehat{c}_m\right)^2$ $C_\alpha(T) = \sum_{m=1}^{|T|}N_mQ_m(T) + \alpha|T|$ $\widehat{\alpha} = \operatorname{argmax}|c_\alpha(T)|$ Split the node into two or more nodes based on $C_\alpha(T)$
End while
**Step 3:** Label each leaf node with its dominant class.
**Step 4:** Return the Tree.

## 4. Experiments

In this section, IMST is discussed using experimental results from a loan audit dataset. The data were derived from the loan records of SMEs provided by a city commercial bank in China in 2020, consisting of 1,428 approved SMEs (Han, 2021).

The algorithmic procedure and main workflow of the experiments are illustrated in Fig. 1. These experiments were all conducted using MATLAB 2024a. Key transformations involving text records are denoted by green squares, while model refinement steps are represented by yellow squares.

These SMEs were categorized into three classes based on quarterly audit results: less attention, normal attention, and more attention (labeled 1, 2, and 3, respectively, in our experiments). All samples were initially used for model training, and a stratified 80 % subset was selected according to the categorical distribution for testing. The distribution of categorical variables in both the training and test datasets is presented in Table 3.

Five representative financial variables—current assets/total assets (CA), retained earnings/total assets (RA), net profit/total assets (NA), equity/total liabilities (EL), and operating income/total assets (OA)—were selected according to the classic Z-score model for credit evaluation. Descriptive statistics for these variables are presented in Table 4.

Four representative categorical variables were also selected, including industry category (Industry), loan term (Term), guarantee type (Guarantee), and credit history (Credit), which were collectively denoted as set N. The frequency distributions of these variables are presented in Table 5.

The correlation coefficients among all numerical variables are presented in Fig. 2. These variables exhibited high intercorrelation; consequently, constructing a decision tree based on a single variable could introduce significant bias.

As credit officers conducted quarterly surveys on these enterprises, the survey texts from the 1,428 SMEs served as the textual data for our experiments. We removed almost all function words, such as articles, prepositions, conjunctions, auxiliary verbs, and interjections, and then constructed the corpus by performing text annotation and word frequency analysis. The frequency and cumulative frequency of the main entities are presented in Fig. 3. This figure presents the 20 most frequent entities extracted from the quarterly survey texts, which formed the foundation for the subsequent text transformation. Attributes were annotated on these entities, and the associated linguistic elements—particularly verbs, adverbs, and adjectives—were identified. These attributes were then formalized as fuzzy sets, enabling the conversion of each survey text into the corresponding fuzzy matrix. Further details regarding the corpus can be found in Han, Liu, Qiang, & Zhang (2023).

**Table 3**
Categorical variable distribution in the training and test datasets.

| Dataset | Categorical-1 | Categorical-2 | Categorical-3 |
|---|---|---|---|
| Training | 35 (13.21 %) | 210 (79.25 %) | 20 (7.55 %) |
| Testing | 176 (15.40 %) | 903 (79.00 %) | 64 (5.60 %) |
| Whole set | 188 (13.17 %) | 1156 (80.95 %) | 84 (5.88 %) |

Note: Figures in parentheses represent the proportions.

**Table 4**
Basic statistics of numerical financial variables.

| Variable | Mean (%) | Std | CV |
|---|---|---|---|
| CA | 21.34 | 19.65 | 92.08 |
| RA | 17.28 | 7.59 | 43.92 |
| NA | 15.38 | 12.37 | 80.43 |
| EL | 64.53 | 20.63 | 31.97 |
| OA | 82.36 | 22.57 | 27.40 |

### 4.1. IMST

Herein, the entries of matrices U and V were initialized with random values drawn from the uniform distribution over the interval (0, 1), and the iterative computation described in Equation (6) was carried out until convergence was achieved. Due to the absence of semantic correspondence in the textual representation of decimal numbers, all numerical values in U and V were rounded to ensure consistency and interpretability. Subsequently, the non-zero eigenvectors were selected, and the corresponding set U was computed. Matrix U was then utilized to construct the IMST, while the transposed matrix V' is presented in Table 6.

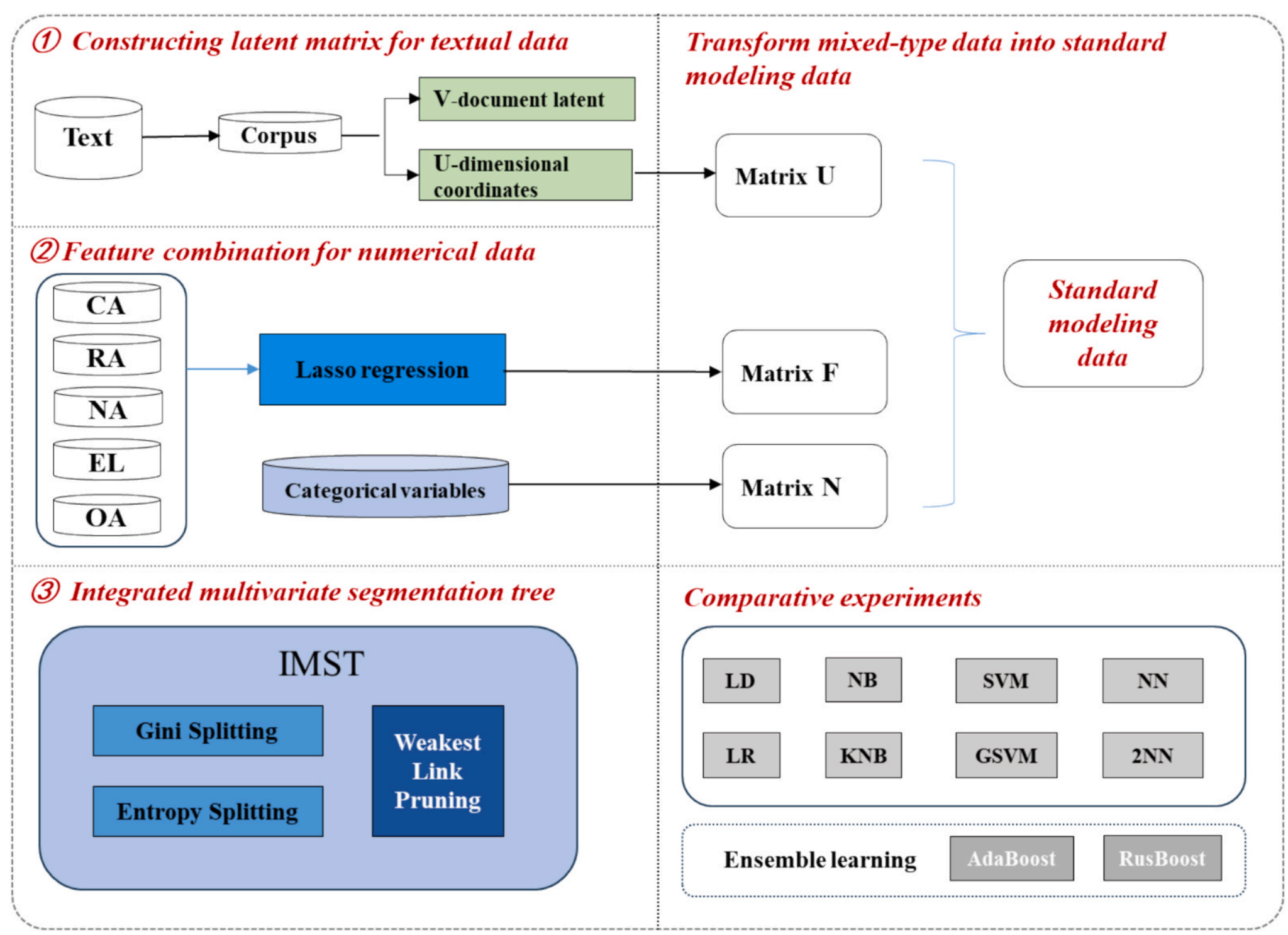

**Fig. 1.** Algorithm and workflow of the experiments.

**Table 5**
Frequencies of categorical and nominal variables.

| Variable | Value | Frequency | Percentage |
|---|---|---|---|
| Industry category | 1 | 209 | 14.63585434 |
| | 2 | 165 | 11.55462185 |
| | 3 | 171 | 11.97478992 |
| | 4 | 206 | 14.42577031 |
| | 5 | 203 | 14.21568627 |
| | 6 | 269 | 18.83753501 |
| Loan term | 1 | 275 | 19.25770308 |
| | 2 | 690 | 48.31932773 |
| | 3 | 152 | 10.6442577 |
| | 4 | 311 | 21.77871148 |
| Guarantee | 0 | 359 | 25.14005602 |
| | 1 | 425 | 29.76190476 |
| | 2 | 644 | 45.09803922 |
| Credit performance | 1 | 318 | 22.26890756 |
| | 2 | 421 | 29.48179272 |
| | 3 | 378 | 26.47058824 |
| | 4 | 311 | 21.77871148 |

To obtain set F, the mean squared error (MSE) of the 10-fold cross validation was computed using Lasso regression with all numerical variables on the entire dataset, as illustrated in Fig. 4. The green circle and dotted line indicate the lambda value corresponding to the minimum cross-validation error, whereas the blue circle and dotted line mark the value associated with the smallest standard error within one standard deviation of the minimum error. Given the small difference between these two values, lambda1SE was selected as the optimal tuning parameter.

The coefficients corresponding to different lambda values are presented in Fig. 5. These results indicated that three of these variable coefficients remained non-zero at lambda1SE.

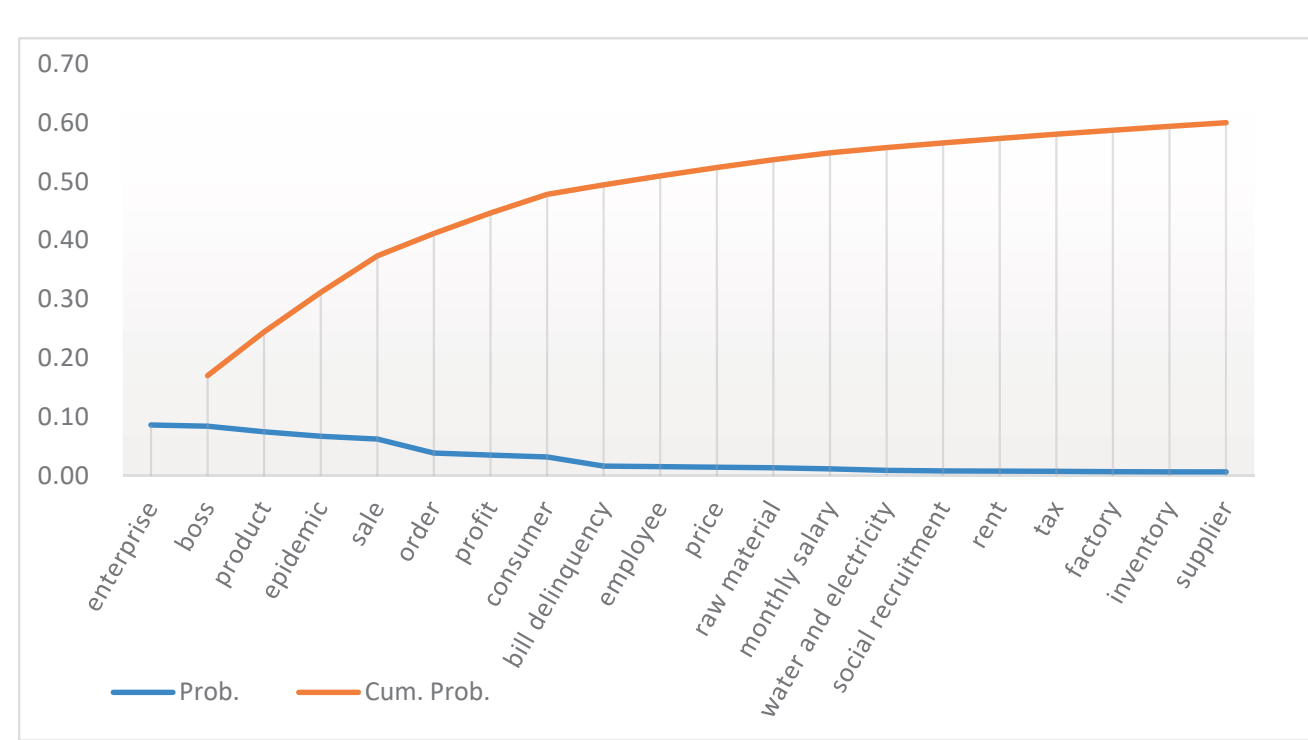

**Fig. 3.** Frequency and cumulative frequency of main entities.

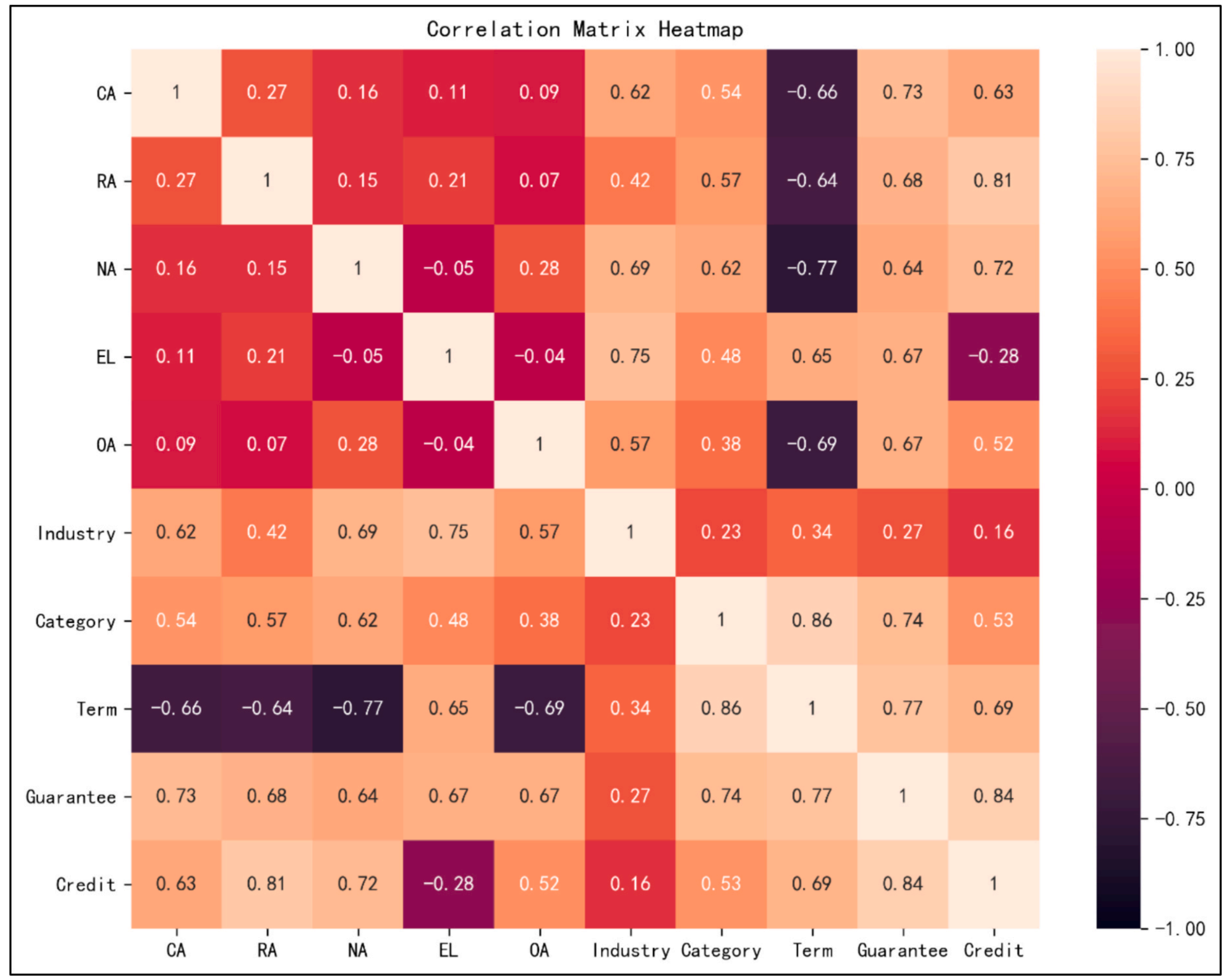

**Fig. 2.** Heatmap of correlation matrix of all numerical variables.

**Table 6**
Latent matrix V' of the original texts.

| | Lat. 1 | Lat. 2 | Lat. 3 | Lat. 4 | Lat. 5 | Lat. 6 |
|---|---|---|---|---|---|---|
| enterprise | 3 | 2 | 1 | 2 | 1 | 0 |
| boss | 3 | 2 | 1 | 1 | 0 | 0 |
| product | 2 | 1 | 1 | 2 | 1 | 0 |
| epidemic | 2 | 2 | 1 | 2 | 0 | 1 |
| sale | 2 | 2 | 1 | 1 | 0 | 0 |
| order | 1 | 1 | 2 | 1 | 0 | 1 |
| profit | 1 | 2 | 1 | 0 | 1 | 1 |
| consumer | 1 | 1 | 2 | 1 | 0 | 1 |
| bill delinquency | 1 | 0 | 2 | 2 | 1 | 0 |
| employee | 1 | 1 | 0 | 1 | 2 | 0 |
| price | 0 | 1 | 1 | 2 | 1 | 0 |
| raw material | 0 | 2 | 1 | 1 | 0 | 1 |
| monthly salary | 0 | 1 | 0 | 2 | 2 | 0 |
| water and electricity | 0 | 0 | 2 | 1 | 2 | 0 |
| social recruitment | 0 | 1 | 0 | 1 | 1 | 2 |
| rent | 0 | 1 | 1 | 0 | 1 | 1 |
| tax | 0 | 1 | 1 | 1 | 1 | 0 |
| factory | 0 | 1 | 0 | 1 | 1 | 1 |
| inventory | 0 | 1 | 1 | 1 | 1 | 0 |
| supplier | 0 | 1 | 1 | 0 | 1 | 1 |

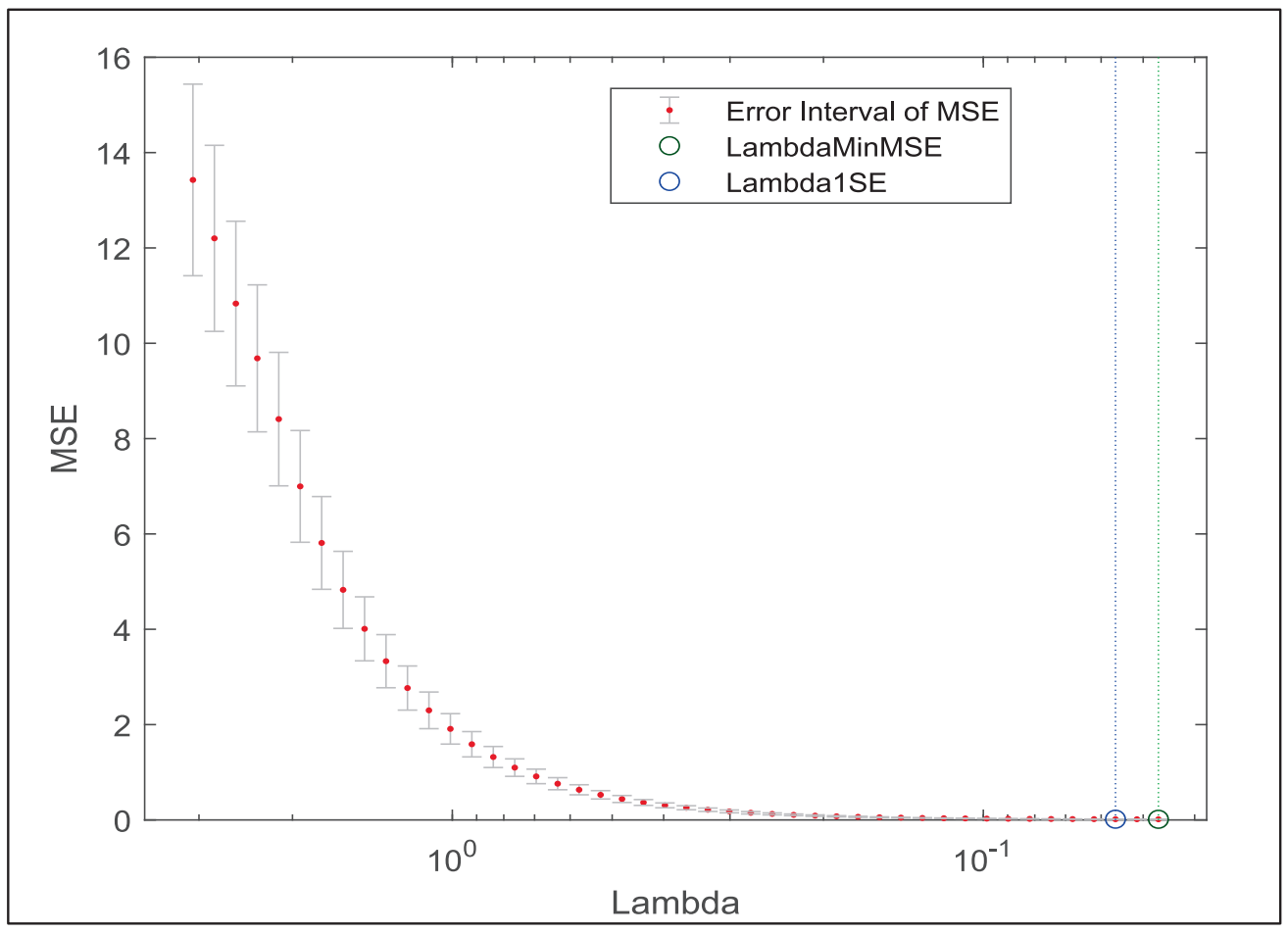

**Fig. 4.** MSE with lambda of 10-fold cross validation.

Therefore, the Lasso regression equation for the numerical variables can be expressed as Eq. (13), where f denotes the estimated value of y. However, in IMST, only the corresponding features had to be selected; neither f nor y was used in the model construction.

$$\widehat{y} = 0.0278 \times CA + 0.0021 \times RA + 0.0435 \times NA$$
$$f = \begin{cases} 1 \ 15\% \text{ of the smallest } \widehat{y} \\ 2 \ 15\% - 95\% \text{ of the } \widehat{y} \\ 3 \text{ Others} \end{cases} \tag{13}$$

Subsequently, the 11 variables—including matrix U, set F, and set N—were integrated to construct the IMST. In this process, the continuous variable f was segmented and categorized into four groups based on its distribution percentiles. Using the weakest-link pruning penalty, both the Gini index and entropy-based splitting criteria yielded structurally similar optimal trees (Fig. 6). The ROC curve, presented in Fig. 7, indicates that the IMST demonstrated strong predictive performance.

When 80 % of the samples were allocated to the test set, the IMST achieved an overall accuracy of approximately 88.9 %, and the corresponding confusion matrix is presented in Fig. 8.

To enable a more rigorous evaluation of method effectiveness, we re-incorporated the 11 variables derived from matrix U, set F, and set N. 70 % of the data samples were assigned to the training set and 30 % to the validation set, with a 5-fold cross-validation procedure subsequently applied. Within this nested cross-validation framework, performance metrics were calculated by averaging results across the five validation folds, ensuring strict separation between evaluation sets and all stages of model development—including hyperparameter tuning and model training—thus minimizing the risk of performance overestimation due to data leakage or distribution bias. The validation results demonstrate that IMST achieved an average accuracy of 76.53 %, with corresponding ROC curves and decision boundaries illustrated in Fig. 9. As shown in Fig. 9, IMST demonstrates superior learning performance in category L, while exhibiting relatively weaker performance in category M.

### 4.2. Baseline model

A decision tree for univariate segmentation was constructed using the same dataset as the baseline model—namely, the nine original variables and six latent variables derived from U. Irrespective of whether the Gini index or entropy was used as the splitting criterion, the resulting tree structures remained largely consistent (Fig. 10).

Test experiment was conducted using the same test samples as those employed in IMST. The baseline model achieved an overall accuracy of approximately 87.4 %, and the corresponding confusion matrix is presented in Fig. 11.

Several interesting findings were drawn from the comparison between the baseline model and IMST. First, IMST outperformed the baseline models in terms of hierarchical structure, primarily because of the reduction in variable dimensions achieved through multivariate segmentation, which considerably simplified the model architecture. Second, in terms of performance, both models achieved high accuracy. IMST performed slightly better, particularly in identifying samples labeled “3,” which required more attention. The accuracy of IMST for these samples was ∼ 60.9 %, compared with ∼ 48.4 % for the baseline model.

Furthermore, the test dataset was partitioned into the following five distinct sample subsets for model evaluation:

- Case 1: Category One, 176; Category Two, 903; Category Three, 0.
- Case 2: Category One, 176; Category Two, 0; Category Three, 64.
- Case 3: Category One, 0; Category Two, 903; Category Three, 64.
- Case 4: Category One, 30; Category Two, 30; Category Three, 30.
- Case 5: Category One, 100; Category Two, 500; Category Three, 50.

Cases 1–3 were designed to evaluate model performance under imbalanced sample conditions; Case 4 assessed performance using small samples that deviated from the expected distribution characteristics; Case 5 evaluated performance with relatively large samples that similarly exhibited nonconforming distributional properties. The corresponding results are presented in Table 7.

As shown in Table 7, IMST consistently outperformed the baseline model in terms of accuracy and F1 score across all sample conditions. Therefore, IMST not only effectively addressed the classification challenges posed by heterogeneous data but also demonstrated strong predictive performance and high interpretability.

### 4.3. Comparison with other classifiers

To enable a comparative analysis with IMST, we implemented multiple classification models using the 9 original variables and 6 latent variables derived from matrix U, along with the full set of samples. 70 % of the data was allocated for training and the remaining 30 % for validation, followed by a 5-fold cross-validation procedure.

All experiments were conducted in a consistent software environment using MATLAB R2024a. The hyperparameters of each baseline model were automatically optimized using MATLAB’s built-in Classification Learner App, a standardized tool for model tuning. Given that extensive hyperparameter optimization is not the primary focus of this study, this approach was adopted to ensure fair comparisons and minimize potential implementation bias while maintaining reproducibility. The key parameter settings for each classifier are summarized in Table 8.

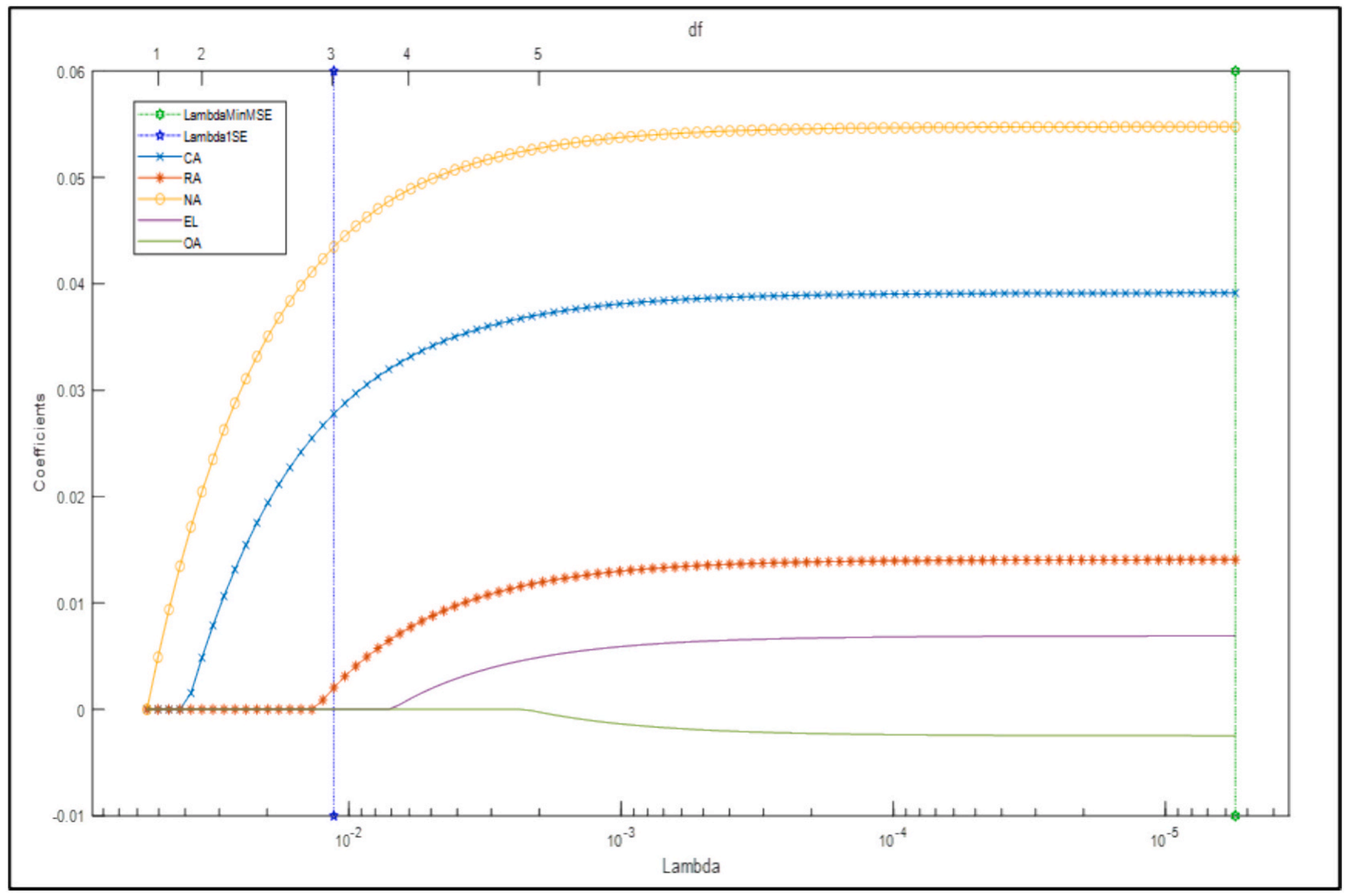

**Fig. 5.** Coefficients with different lambda values.

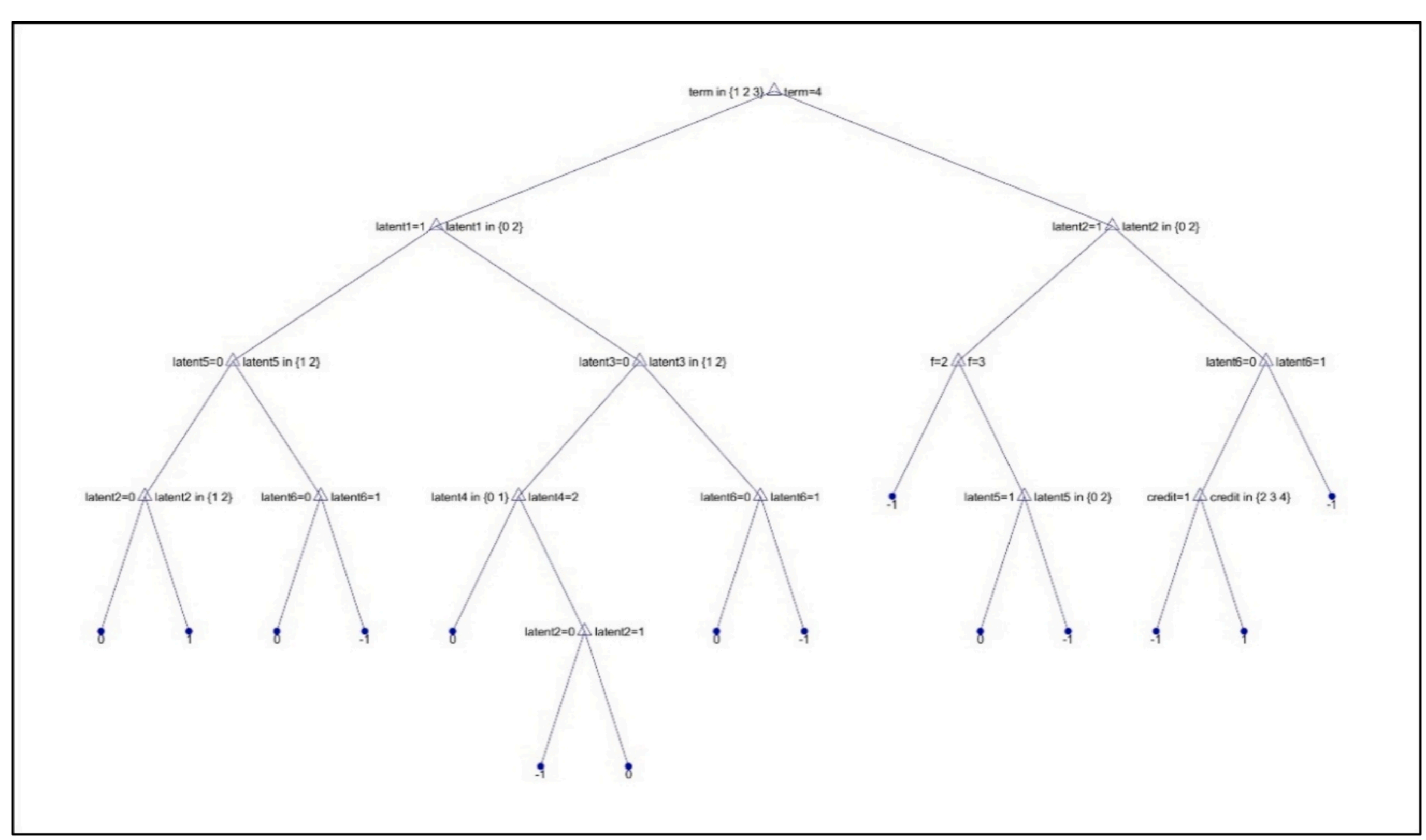

**Fig. 6.** IMST representation.

Performance metrics are reported as mean values obtained from 5-fold cross-validation to enhance the reliability of the performance estimates. The results are summarized in Table 9.

Table 9 shows that IMST achieved the highest accuracy. Most classifiers demonstrated good performance, except for neural network classifiers, which performed relatively poorly on this dataset. This might be attributed to the sparsity resulting from the large number of zero values in the latent features of these text records, which decreased the recognition capability of the neural networks.

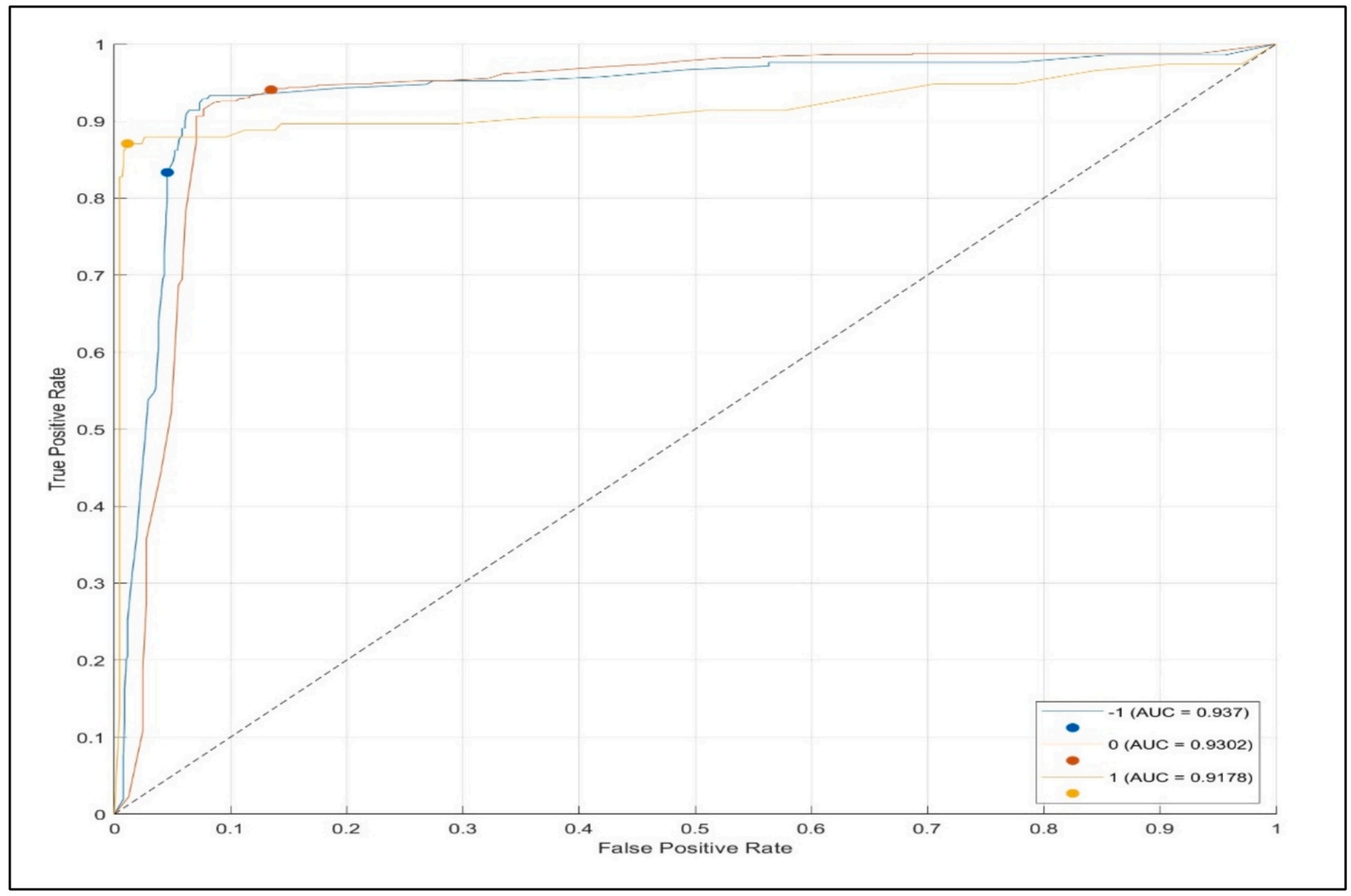

**Fig. 7.** ROC curve of IMST.

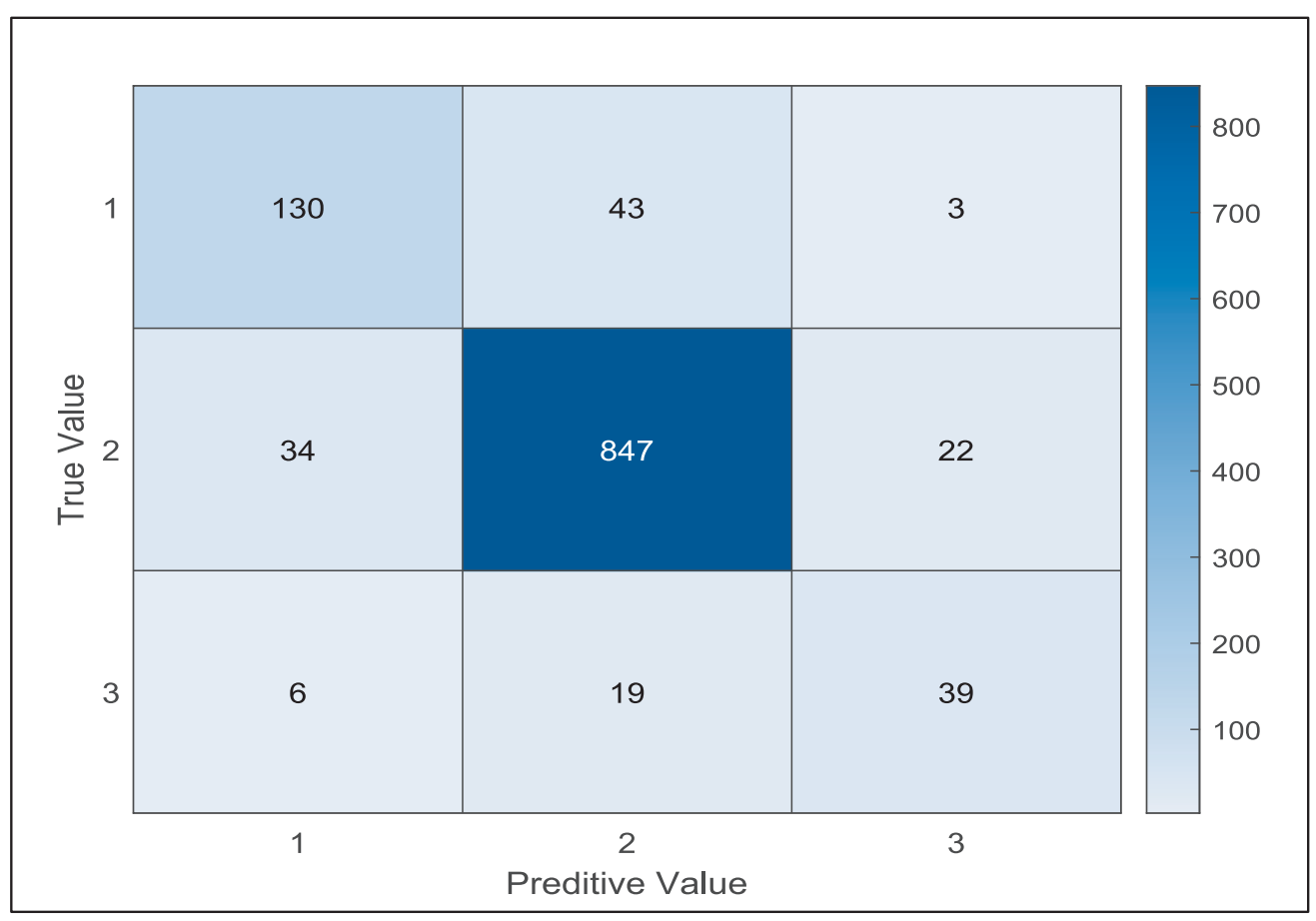

**Fig. 8.** Confusion matrix of IMST.

The accuracy results of model recognition for categories requiring more and less attention are presented in Fig. 12. This figure shows that the naïve Bayes models slightly outperformed the other classifiers in the “more attention” groups. Although the performance of IMST was not optimal on the “more attention” group, it demonstrated considerably better predictive ability on the “less attention” group compared with the other models. Accurately classifying the “less attention” group can substantially reduce survey costs, offering particular advantages for financial institutions such as banks. Therefore, we conclude that IMST represents the most practical and effective approach for real-world deployment.

## 5. Discussion

There are several important considerations regarding IMST.

First, from a methodological standpoint, combining categorical and numerical variables in Lasso regression is technically feasible. However, this approach presents certain challenges. Categorical variables typically exhibit substantially lower variance than numerical variables, often resulting in smaller regression coefficients after Lasso regularization. These attenuated coefficients can negatively affect subsequent decision tree classification, particularly in terms of feature importance and model interpretability. Furthermore, categorical variables are inherently discrete and generally contribute less to the structural complexity of decision tree construction than continuous variables. In our experimental evaluation, applying Lasso regression to both variable types before IMST leads to a considerable decline in the overall model accuracy. This degradation is primarily attributed to the fact that only two out of the four categorical variables yield non-zero coefficients under Lasso regularization. Lasso regression helps not only with variable selection and dimensionality reduction but also in handling continuous features within decision tree models. Therefore, we applied it exclusively to numerical variables before constructing the multivariate segmentation tree.

Second, textual data preprocessing can considerably influence algorithmic performance. Herein, we used Baidu’s word segmentation and part-of-speech tagging tools, which first identified entities and then extracted attributes associated with these entities. However, during processing, we observed limitations in the semantic recognition capability. In particular, the system failed to recognize that semantically equivalent expressions might convey the same underlying meaning. For instance, sentences such as “the boss is not here” and “the boss is on a business trip” were semantically similar yet treated as distinct textual instances by the tool. This discrepancy might introduce bias into the constructed word matrix and training corpus, potentially affecting the generalizability of the model.

Third, the sparsity of the text latent matrix presented a considerable challenge for classification tasks. Although dimensionality reduction was effectively achieved compared with the original text records, the latent matrix still contained a high proportion of zero-valued entries, which reduced data variability and limited feature expressiveness. Although decision tree models were relatively robust to such sparsity because of their intrinsic partitioning mechanisms, other classical

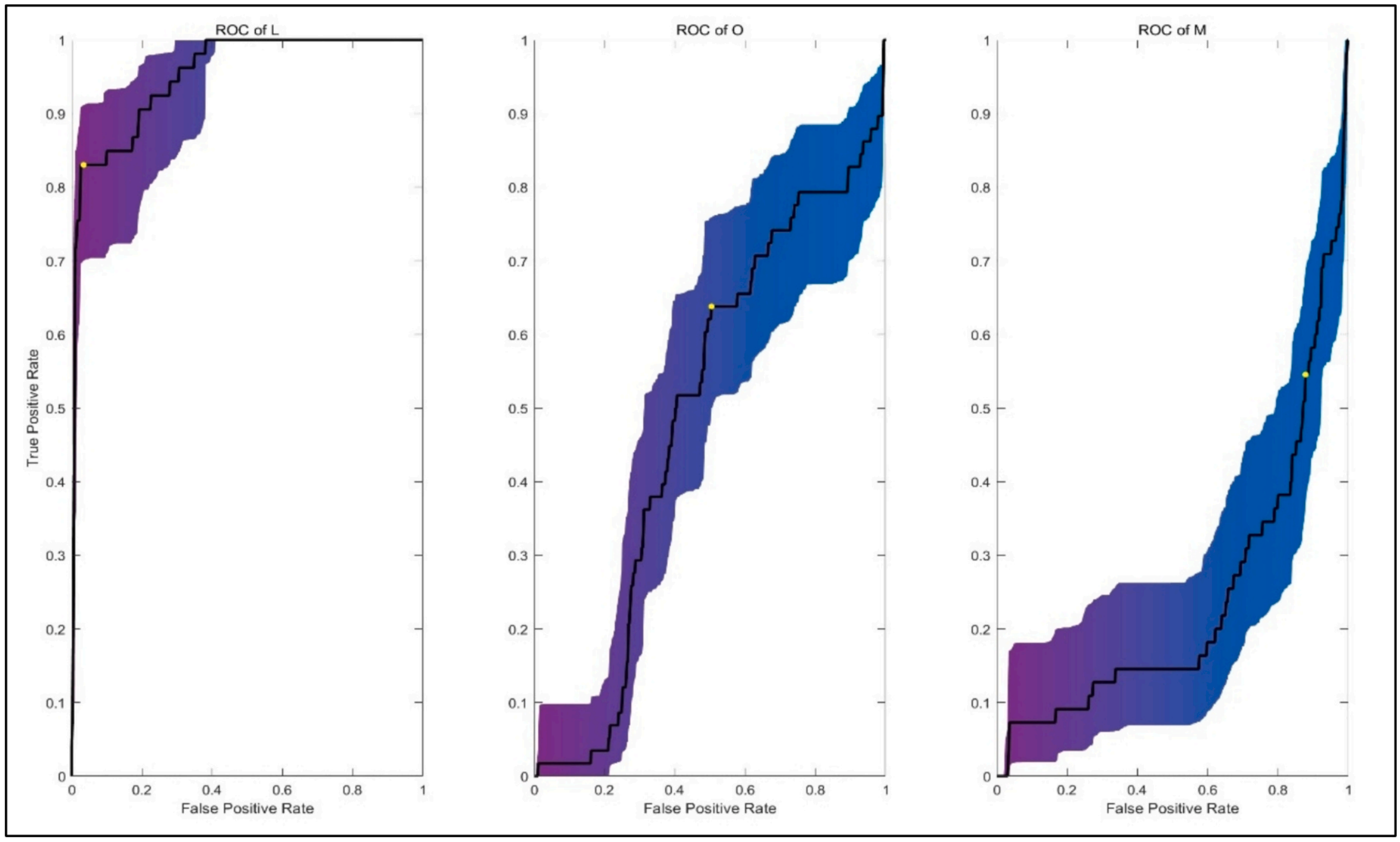

**Fig. 9.** ROC curves and decision boundaries of the validation sets across different categories.

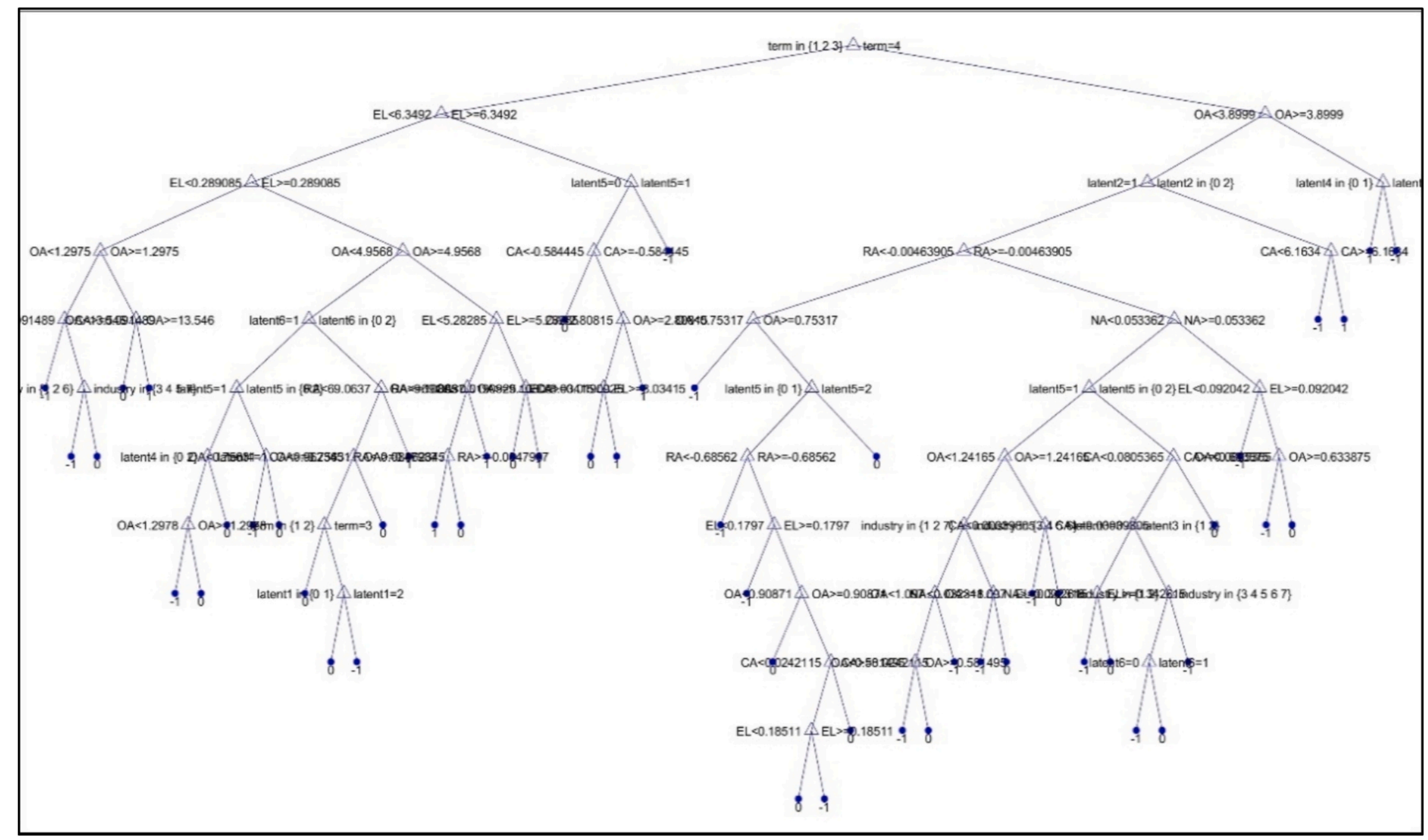

**Fig. 10.** Baseline model representation.

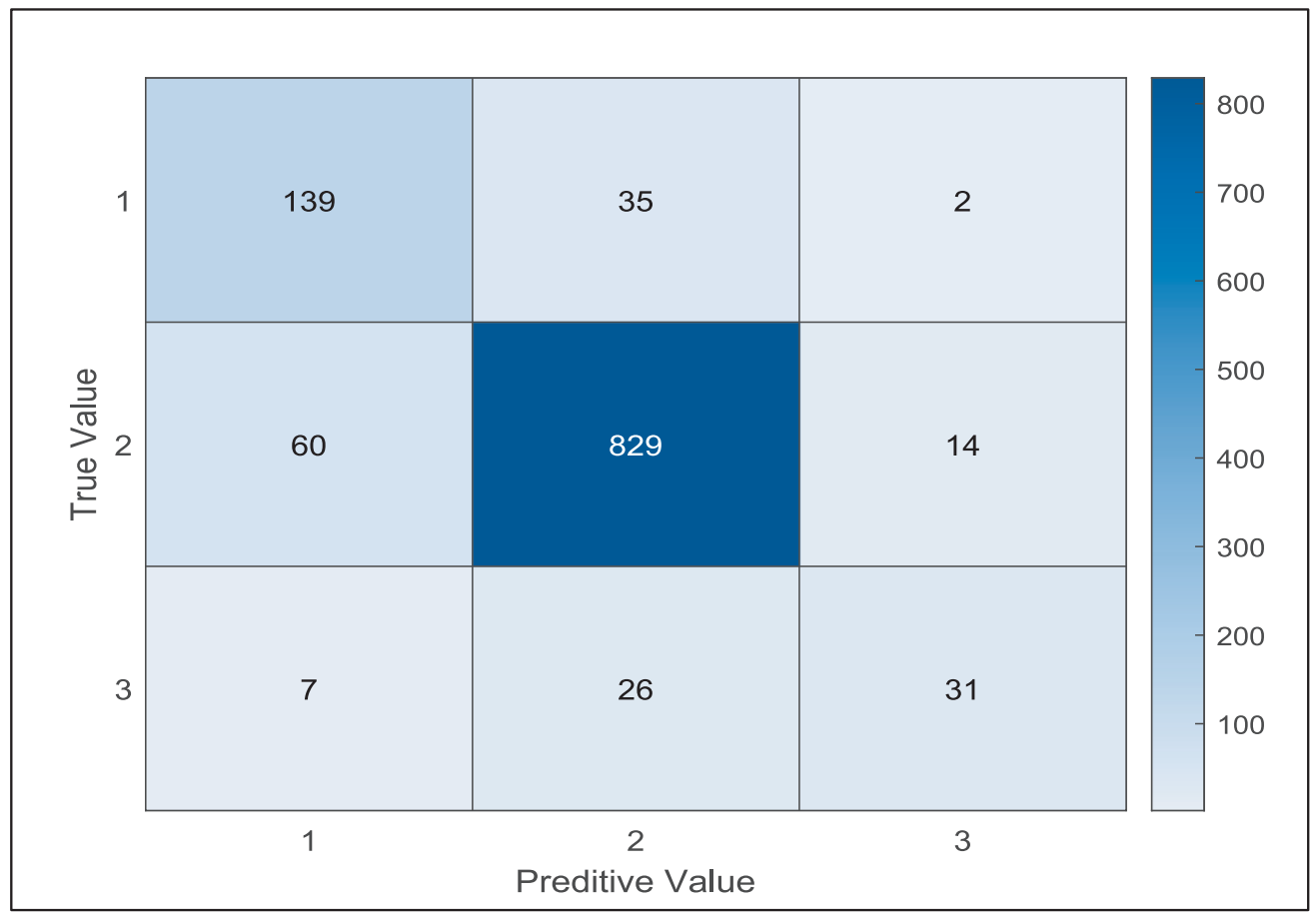

**Fig. 11.** Confusion matrix of baseline model.

**Table 7**
Comparison of model results across five distinct sample subsets.

| Cases | Baseline model | | | | IMST | | | |
|---|---|---|---|---|---|---|---|---|
| | Accu. | Prec. | Rec. | F1 | Accu. | Prec. | Rec. | F1 |
| **Case 1** | 45 % | 18 % | 38 % | 27 % | 51 % | 27 % | 48 % | 37 % |
| **Case 2** | 37 % | 14 % | 33 % | 24 % | 47 % | 43 % | 27 % | 34 % |
| **Case 3** | 52 % | 28 % | 46 % | 34 % | 59 % | 65 % | 36 % | 42 % |
| **Case 4** | 61 % | 35 % | 52 % | 43 % | 64 % | 67 % | 41 % | 48 % |
| **Case 5** | 67 % | 46 % | 58 % | 52 % | 71 % | 68 % | 38 % | 57 % |

Note:
1. Accuracy, recall, and F1-score are all average values across all categories.

**Table 8**
Configuration of different classifiers.

| Classifier (Abbreviations) | Configuration |
|---|---|
| **Linear Discrimination (LD)** | Covariance Structure: Full |
| **Logistic Regression (LR)** | Regularization: Auto<br>Lambda: Auto<br>Beta:0.0001 |
| **Naive Bayes (NB)** | Kernel Function: None<br>Standardized: True |
| **Kernel naïve Bayes (KNB)** | Kernel Function: Gaussian Kernel Function<br>Standardized: True |
| **Linear SVM (LSVM)** | Kernel Function: None<br>Kernel Scale:1<br>Frame Constraint Level: 1<br>Standardize: True |
| **Gaussian kernel SVM (GSVM)** | Kernel Function: Gaussian Kernel Function<br>Kernel Scale:0.83<br>Frame Constraint Level: 1<br>Standardize: True |
| **Neural Network (NN)** | Number of Fully Connected Layers: 1<br>Size of the First Layer: 25<br>Activation Function: ReLU<br>Iteration Limit: 1000<br>Lambda: 0<br>Standardize: True |
| **Double-layer Neural Network (2NN)** | Number of Fully Connected Layers: 2<br>Size of the First Layer: 10<br>Size of the Second Layer: 10<br>Activation Function: ReLU<br>Iteration Limit: 1000<br>Lambda: 0<br>Standardize: True |
| **AdaBoost (AdaB)** | Maximum Number of Splits: 20<br>Number of Learners: 30<br>Learning Rate: 0.1<br>Number of Predictor Variables Used for Sampling: All |
| **RusBoost (RusB)** | Maximum Number of Splits: 20<br>Number of Learners: 30<br>Learning Rate: 0.1<br>Number of Predictor Variables Used for Sampling: All |

**Table 9**
Performance of different classifiers.

| Classifier | Accuracy (%) | Time (s) |
|---|---|---|
| **IMST** | 76.53 | 3.16 |
| **AdaBoost** | 71.42 | 3.24 |
| **RusBoost** | 69.97 | 4.08 |
| **Linear Discrimination** | 72.38 | 1.86 |
| **Logistic Regression** | 70.05 | 3.56 |
| **Naive Bayes** | 68.37 | 4.67 |
| **Kernel naïve Bayes** | 63.53 | 4.62 |
| **Linear SVM** | 67.73 | 3.87 |
| **Gaussian kernel SVM** | 64.39 | 4.86 |
| **Neural Network** | 60.17 | 4.86 |
| **Double-layer Neural Network** | 59.38 | 6.48 |

classifiers, particularly neural networks, were substantially affected, exhibiting notable declines in classification accuracy. This performance degradation is not attributable to inherent deficiencies in the classifiers themselves; rather, it stems from the limitations in the representation process. Therefore, further research is warranted to improve information density in matrix-based representations of textual data.

Finally, the decision to adopt a decision tree classifier for handling heterogeneous data was primarily motivated by its superior interpretability to that of alternative classification methods. Decision trees enabled the direct extraction of human-readable decision rules, facilitating model transparency and supporting domain-specific interpretation. Furthermore, experimental results indicated that, with the exception of neural networks, most classifiers achieved high accuracy when using text latent features. This observation suggested that improving classifier performance through the optimization of text latent feature representations might constitute a promising direction for future research.

## 6. Conclusion

Herein, we proposed a novel method—IMST—for classification tasks involving heterogeneous data. The method transformed textual data into a latent matrix through matrix decomposition, applied Lasso regression to select salient features for constructing multivariate segmentation relationships, and subsequently used weakest-link pruning optimization to build the decision tree. Our approach to generating text latent representations converted unstructured textual information into structured numerical matrices, ensuring compatibility with a broad range of machine learning classifiers. Moreover, the integration of Lasso regression in forming multivariate segmentation criteria not only substantially decreased the model complexity but also effectively managed continuous variables, thereby improving the method's adaptability across various decision tree algorithms. Experimental results based on real-world data from a city commercial bank demonstrated that IMST considerably simplified the decision tree structure, improved computational efficiency, and achieved competitive performance in terms of accuracy and interpretability. These outcomes suggest that IMST represents a promising direction for modeling and analyzing heterogeneous data in practical applications.

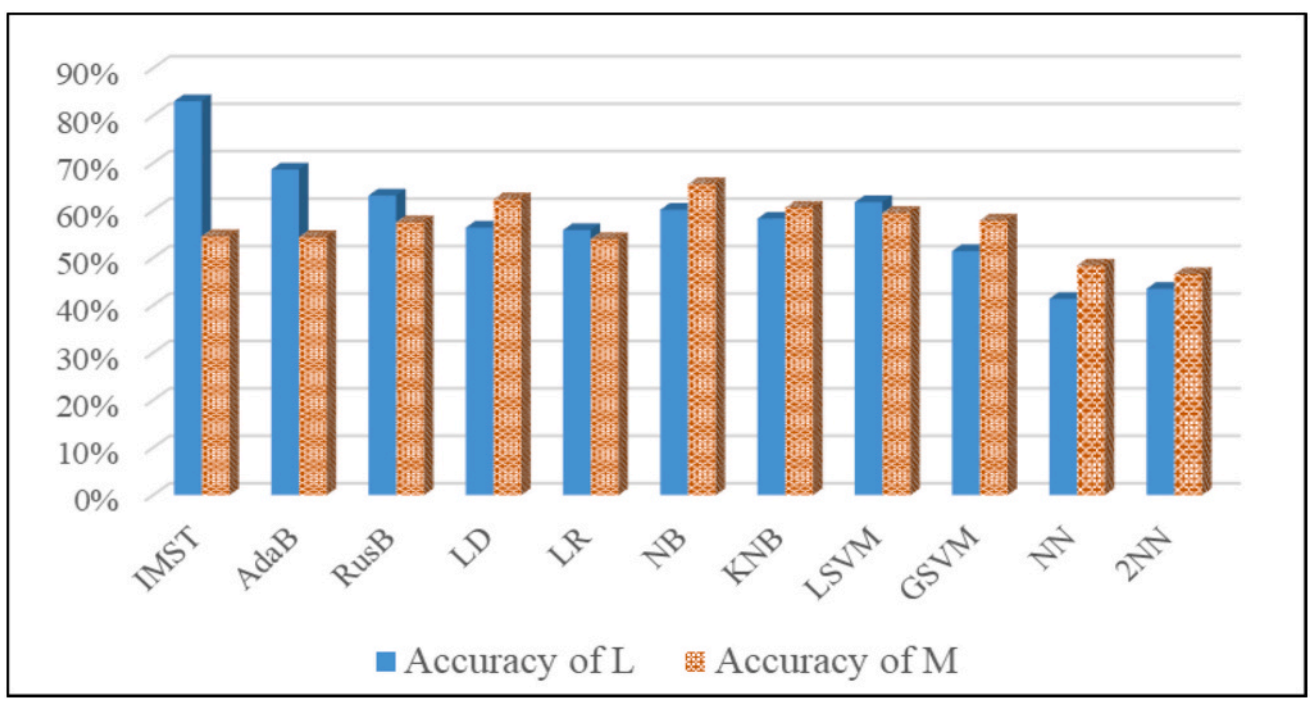

**Fig. 12.** Accuracy of different groups with classifiers.

## CRediT authorship contribution statement

**Lu Han:** Conceptualization, Methodology, Formal analysis, Investigation, Resources, Data curation, Writing – original draft, Writing – review & editing, Visualization, Funding acquisition, Validation. **Xiuying Wang:** Conceptualization, Methodology, Formal analysis, Investigation, Resources, Writing – review & editing, Validation.

## Declaration of competing interest

The authors declare that they have no known competing financial interests or personal relationships that could have appeared to influence the work reported in this paper.

## Acknowledgment

The work was supported by the National Natural Science Foundation of China (Grant No. 72101279), Visiting Scholar Grant Program of China Scholarship Council for Han (No. 202406490006), and the Fundamental Research Funds for the Central Universities (No. JYXZ2407).

## Data availability

Data will be made available on request.

## References

Abolhosseini, S., Khorashadizadeh, M., Chahkandi, M., & Golalizadeh, M. (2024). A modified ID3 decision tree algorithm based on cumulative residual entropy. *Expert Systems with Applications, 255*.

Airlangga, G., & Liu, A. (2025). A hybrid gradient boosting and neural network model for predicting urban happiness: Integrating ensemble learning with deep representation for enhanced accuracy. *Machine Learning and Knowledge Extraction, 7*(1), 4.

Arifuzzaman, M., Hasan, M. R., Toma, T. J., Hassan, S. B., & Paul, A. K. (2023). An advanced decision tree-based deep neural network in nonlinear data classification. *Technologies, 11*(1), 24.

Blockeel, H., Devos, L., Frénay, B., Nanfack, G., & Nijssen, S. (2023). Decision trees: From efficient prediction to responsible AI. *Frontiers in Artificial Intelligence, 6*, Article 1124553.

Browne, R. P., & McNicholas, P. D. (2012). Model-based clustering, classification, and discriminant analysis of data with mixed type. *Journal of Statistical Planning and Inference, 142*(11), 2976–2984.

Carrasco, L., Urrutia, F., Abeliuk, A., 2025. Zero-shot decision tree construction via large language models: arXiv. http://arxiv.org/abs/2501.16247.

Charbuty, B., & Abdulazeez, A. (2021). Classification based on decision tree algorithm for machine learning. *Journal of Applied Science and Technology Trends, 2*(01), 20–28.

Emami, S., & Martínez-Muñoz, G. (2025). Condensed-gradient boosting. *International Journal of Machine Learning and Cybernetics, 16*(1), 687–701.

Gao, J., Li, P., Chen, Z., & Zhang, J. (2020). A survey on deep learning for multimodal data fusion. *Neural Computation, 32*(5), 829–864.

Gupta, S., & Kishan, B. (2025). A performance-driven hybrid text-image classification model for multimodal data. *Scientific Reports, 15*(1), 11598.

Gutierrez-Gomez, L., Petry, F., & Khadraoui, D. (2020). A comparison framework of machine learning algorithms for mixed-type variables datasets: A case study on tire-performances prediction. *IEEE Access, 8*, 214902–214914.

Han, L., Li, W., & Su, Z. (2019). An assertive reasoning method for emergency response management based on knowledge elements C4.5 decision tree. *Expert Systems with Applications, 122*, 65–74.

Han, L., Liu, Z., Qiang, J., & Zhang, Z. (2023). Fuzzy clustering analysis for the loan audit short texts. *Knowledge and Information Systems, 65*(12), 5331–5351.

Han, L. (2021). Credit investigation data of SMEs provided by a city commercial bank in China in 2020. https://www.researchgate.net/publication/390694764_Credit_investigation_of_SMEs.

Hsu, C., Tsao, W., Chang, A., & Chang, C. (2021). Analyzing mixed-type data by using word embedding for handling categorical features. *Intelligent Data Analysis, 25*(6), 1349–1368.

Jiang, C., Yin, C., Tang, Q., & Wang, Z. (2023). The value of official website information in the credit risk evaluation of SMEs. *Journal of Business Research, 169*, Article 114290.

Khalaf, O., Garcia, S., & Ben Ishak, A. (2025). Generalised entropies for decision trees in classification under monotonicity constraints. *Expert Systems, 42*(4).

Khan, A. A., Chaudhari, O., & Chandra, R. (2024). A review of ensemble learning and data augmentation models for class imbalanced problems: Combination, implementation and evaluation. *Expert Systems with Applications, 244*.

Kim, K., & Hong, J. (2017). A hybrid decision tree algorithm for mixed numeric and categorical data in regression analysis. *Pattern Recognition Letters, 98*, 39–45.

Lee, J. Y., Yang, J., & Anderson, E. T. (2024). Using grocery data for credit decisions. *Management Science*.

Lee, Y., Yen, S., Jiang, W., Chen, J., & Chang, C. (2025). Illuminating the black box: An interpretable machine learning based on ensemble trees. *Expert Systems with Applications, 272*.

Li, S., Tang, H. (2024). Multimodal Alignment and Fusion: A Survey: arXiv. http://arxiv.org/abs/2411.17040.

Li, Y. C., Wang, L., Law, J. N., Murali, T. M., Pandey, G., & Lengauer, T. (2022). Integrating multimodal data through interpretable heterogeneous ensembles. *Bioinformatics Advances, 2*(1), Article vbac65.

Li, Y., Herrera-Viedma, E., Kou, G., & Morente-Molinera, J. A. (2023). Z-number-valued rule-based decision trees. *Information Sciences, 643*.

Lu, Y., Chen, F., Wang, Y., & Lu, C. (2016). Discovering anomalies on mixed-type data using a generalized student- based approach. *IEEE Transactions on Knowledge and Data Engineering, 28*(10), 2582–2595.

Mantas, C. J., & Abellán, J. (2014). Credal-C4.5: Decision tree based on imprecise probabilities to classify noisy data. *Expert Systems with Applications, 41*(10), 4625–4637.

Mantas, C. J., Abellán, J., & Castellano, J. G. (2016). Analysis of Credal-C4.5 for classification in noisy domains. *Expert Systems with Applications, 61*, 314–326.

Martin, N. (2013). Assessing scorecard performance: A literature review and classification. *Expert Systems with Applications, 40*(16), 6340–6350.

Mienye, I. D., & Jere, N. (2024). A survey of decision trees: Concepts, algorithms, and applications. *IEEE Access, 12*, 86716–86727.

Moral-García, S., Mantas, C. J., Castellano, J. G., Benítez, M. D., & Abellán, J. (2020). Bagging of credal decision trees for imprecise classification. *Expert Systems with Applications, 141*, Article 112944.

Paim, A. M., & Enembreck, F. (2025). Adaptive random tree ensemble for evolving data stream classification. *Knowledge-Based Systems, 309*, Article 112830.

Stahlschmidt, S. R., Ulfenborg, B., & Synnergren, J. (2022). Multimodal deep learning for biomedical data fusion: A review. *Briefings in Bioinformatics, 23*(2), Article bbab569.

Tao, Q., Li, Z., Xu, J., Xie, N., Wang, S., & Suykens, J. A. K. (2021). Learning with continuous piecewise linear decision trees. *Expert Systems with Applications, 168*.

Thirunavukkarasu, M., & Jamal, N. (2025). Hybrid machine learning classifier models for kidney disease detection. *Informatica, 49*(7).

Xia, M., Wang, Z., Lan, X., Liu, W., Wu, J. (2025). Confidence-driven under-sampling decision forest for imbalanced credit scoring. *Engineering Applications of Artificial Intelligence*, 147.

Xu, W., & Tian, Z. (2025). Feature selection and information fusion based on preference ranking organization method in interval-valued multi-source decision-making information systems. *Information Sciences, 700*, Article 121860.

Yang, T., Yan, F., Qiao, F., Wang, J., & Qian, Y. (2025). Fusing monotonic decision tree based on related family. *IEEE Transactions on Knowledge and Data Engineering, 37*(2), 670–684.

Zhang, X., & Yu, L. (2024). Consumer credit risk assessment: A review from the state-of-the-art classification algorithms, data traits, and learning methods. *Expert Systems with Applications, 237*, Article 121484.

Zhang, X., & Jiang, S. (2012). A splitting criteria based on similarity in decision tree learning. *Journal of Software, 7*(8), 1775–1782.

Zhao, F., Zhang, C., & Geng, B. (2024). Deep multimodal data fusion. *ACM Computing Surveys, 56*(9), 1–36.

Zhou, H., Zhang, J., Zhou, Y., Guo, X., & Ma, Y. (2021). A feature selection algorithm of decision tree based on feature weight. *Expert Systems with Applications, 164*.